\documentclass[balance,upint,subscriptcorrection,varvw,mathalfa=cal=boondoxo,spanish,french,vietnamese,russian,greek,pdf-a,colorlinks]{asmeconf}

\usepackage{multirow}
\usepackage{multicol}
\usepackage{gensymb}
\usepackage{tabularx}
\usepackage{booktabs,multirow,siunitx}
\hypersetup{%
	pdfauthor={John H. Lienhard},									  
	pdftitle={ASME Conference Paper LaTeX Template},                  
	pdfkeywords={ASME conference paper, LaTeX template, BibTeX style},
	pdfsubject = {Describes the asmeconf LaTeX template},			  
	pdflicenseurl={https://ctan.org/pkg/asmeconf},
}

\begin{document}




\title{Few-Shot VLM-Based G-Code and HMI Verification in CNC Machining}

\SetAuthors{%
	Yasaman Hashem Pour\affil{}\CorrespondingAuthor{yhashemp@mtu.edu}, 
    Nazanin Mahjourian, 
	Vinh Nguyen
}

\SetAffiliation{}{Department of Mechanical and Aerospace Engineering,\\
Michigan Technological University, Houghton, MI 49931}




\maketitle





\begin{abstract}

Manual generation of G-code is important for learning the operation of CNC machines. Prior work in G-code verification uses Large-Language Models (LLMs), which primarily examine errors in the written programming. However, CNC machining requires extensive use and knowledge of the Human-Machine Interface (HMI), which displays machine status and errors. LLMs currently lack the capability to leverage knowledge of HMIs due to their inability to access the vision modality. This paper proposes a few-shot VLM-based verification approach that simultaneously evaluates the G-code and the HMI display for errors and safety status. The input dataset includes paired G-code text and associated HMI screenshots from a 15-slant-PRO lathe, including both correct and error-prone cases. To enable few-shot learning, the VLM is provided with a structured JSON schema based on prior heuristic knowledge. After determining the prompts, instances of G-code and HMI that either contain errors or are error free are used as few-shot examples to guide the VLM. The model was then evaluated in comparison to a zero-shot VLM through multiple scenarios of incorrect G-code and HMI errors with respect to per-slot accuracy. The VLM showed that few-shot prompting led to overall enhancement of detecting HMI errors and discrepancies with the G-code for more comprehensive debugging. Therefore, the proposed framework was demonstrated to be suitable for verification of manually generated G-code that is typically developed in CNC training.

\end{abstract}






\section{Introduction}


While learning to write G-code is essential for understanding CNC machining, it can also be complex, error-prone, and time-consuming \cite{Kamran2021}. This is because G-code is a low-level language that also requires knowledge of the environment in which the code is hosted, such as the machine's characteristics, process conditions, and the human-machine interface (HMI) display. CNC errors are typically caused by manual mistakes, user-defined toolpath routines, and high interpolation sequence codes that require user correction\cite{Schmitt2024,Leary2023}. Even with advanced G-code generation software, obtaining accurate and reliable G-code requires deep technical knowledge and manual debugging, including thorough verification \cite{abdelaal2025gllmselfcorrectivegcodegeneration,jignasu2023foundationalaimodelsadditive}. In practice, manual G-code debugging is vulnerable to extraneous variables, including human error, incorrect coordinate systems, and missing tool calls. This is because troubleshooting these errors demands spatial and technical reasoning \cite{jignasu2023foundationalaimodelsadditive}. Therefore, careful verification of the G-code is key to avoiding costly errors and facilitating safe and precise operation. Intelligent verification systems can help analyze and detect potential G-code problems before machining begins \cite{BADINI2023278, 11021219}.

In recent research, many models including Large Language Models (LLMs) have been applied to analyze and correct G-code automatically \cite{Jignasu2024,BADINI2023278,SHARMA2024111934,deng2024vmadvisualenhancedmultimodallarge}. Various studies show considerable potential for LLMs interpreting and optimizing G-code within manufacturing applications. By leveraging pattern recognition and contextual understanding of programming syntax, these models can analyze G-code and identify structural and logic errors \cite{He2025}. Furthermore, the LLM approaches are also capable of self-correction and anomaly detection even in zero-shot settings, which leads to enhanced code reliability and minimizes reference errors \cite{10.1145/3690407.3690448}. Such models have been employed to generate optimized G-code. The optimized G-code can improve mechanical performance and manufacturing reliability in additive manufacturing \cite{Pilch2025}. More recently, the developing body of literature has shifted toward exploring explainability and interpretability. This approach can help to detect manufacturing features and enhance process understanding and robustness through chain-of-thought prompting \cite{XuanLiu2025}.
As a result of these strengths, LLM-based frameworks are capable of interpreting and analyzing low-level G-code structures and handling corrective tasks. 
However, despite their effectiveness in text-based reasoning, the implementation of LLMs in CNC machining faces significant limitations. Current LLM systems are mainly based on tokenized text and cannot process visual or sensory feedback that characterizes the machining environment \cite{Jignasu2024, XuanLiu2025}. In addition, many systems simulate machine signals instead of monitoring actual machine data, causing them to struggle with adapting to tool conditions and variations in cutting speed and load \cite{Pilch2025}. Hence, LLMs are incapable of aligning G-code with actual machine behavior since they operate offline without interfacing with key monitoring devices, alarms, feedrates, and spindle activity \cite{He2025}. Consequently, verification using LLMs is restricted to syntax correction and G-code optimization \cite{XuanLiu2025}.
 
To mitigate the absence of contextual awareness in LLMs, a number of studies have integrated multimodal data. Vision-based systems 
have been adopted to detect process anomalies, tool damage, and quality deviations by integrating visual or sensory input\cite{LINS2020,Sawangsri2022,Zhang202}. Other methods combine machine vision with digital twin and virtual environments to allow closed-loop control and real-time process feedback \cite{DANG2025106213,Zhong2025}.
In addition to sensor-based monitoring, a few studies investigated the use of HMI data to connect to machine sensor outputs. Some studies rely on sensor-derived parameters, including spindle load, torque, and axis motion data, to evaluate the machining state. In contrast, others integrate user interaction through text-based instructions or voice-guided systems for indirect monitoring \cite{BATTINI2022107881, Stateczny2021,Gunawan2025}.

Existing studies on adaptive CNC interfaces have investigated various approaches for using HMI displays. They aimed to build unified intelligent platforms that operate across multiple CNC machine types and provide improved NC data handling. However, these systems still fall short of integrating G-code with HMI feedback in a meaningful way \cite{Park2024}. Other work has achieved partial integration by routing sorted coordinate-position G-code through the CNC controller while simultaneously managing extrusion and temperature-related G-code for manufacturing processes\cite{Lin2023FDM_CNC}. Additionally, some Tool Condition Monitoring studies connect to NC Blocks to extract specific G-code blocks or align tool-life strategies with machining commands, but these integrations are complicated and limited\cite{Mohamed2022}. These HMI systems visualize machining parameters, tool status, and system alarms to provide a partial understanding of machine state through observable indicators \cite{Sawangsri2022,LINS2020}. CNC machining depends on system feedback presented predominantly in HMI dashboards, where information is represented through text and symbols. However, no current study explicitly links G-code text with HMI data to verify machining G-code or the status indicators displayed on the HMI.

This study introduces a multimodal VLM-based framework that jointly examines  G-code syntax alongside visual information extracted from HMI panels to detect discrepancies, incomplete code segments, and machine-state mismatches. A paired dataset was constructed from G-code programs and their corresponding HMI screenshots to enable the model to infer associations between program commands and virtual cues. Through this integration, the framework can detect both textual and visual inconsistencies that provide a more comprehensive basis for debugging and safety assurance during manual G-code development. Additionally, a JSON-based structured schema was developed to capture heuristic machining knowledge, including expected tool states, feed rates, and spindle behaviors to inform model reasoning. Furthermore, few-shot prompting was employed to compare the proposed framework with a zero-shot configuration, revealing that the model adapts more effectively to diverse machining contexts when provided with a few examples.

First, the methodology of the VLM framework is presented. Then, the experimental setup for evaluation is described. The results are then presented, followed by concluding remarks.


\section{Method}
\label{sec:method}
\begin{figure}[h!]
    \centering
    \includegraphics[width=0.98\linewidth, keepaspectratio]{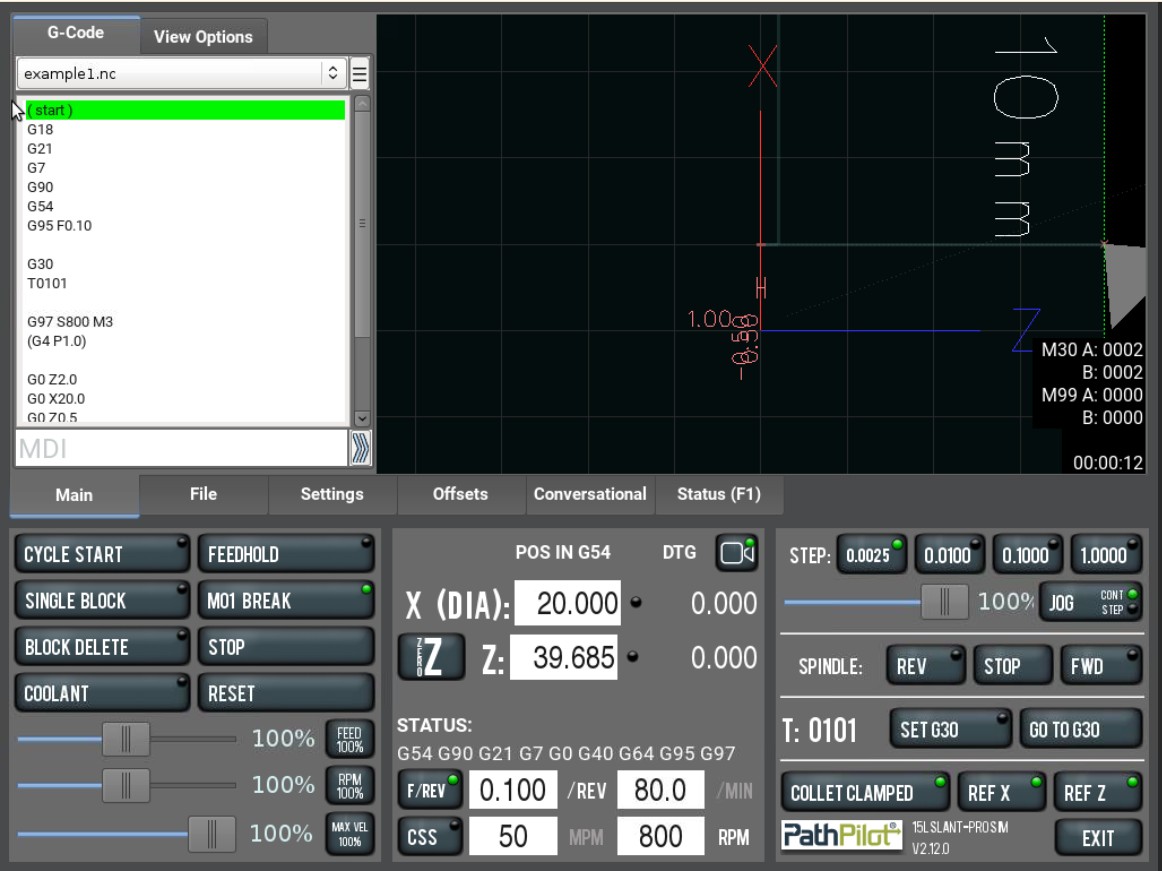}
    \caption{Example of the HMI interface used for verification. The screenshot shows the control-panel indicators (collet clamp, REF X, and REF Z) that the model analyzes to assess machine-state consistency with the input G-code.}
    \label{fig:HMI}
\end{figure}

\begin{figure*}[htb!]
    \centering
    \includegraphics[width=\linewidth, keepaspectratio]{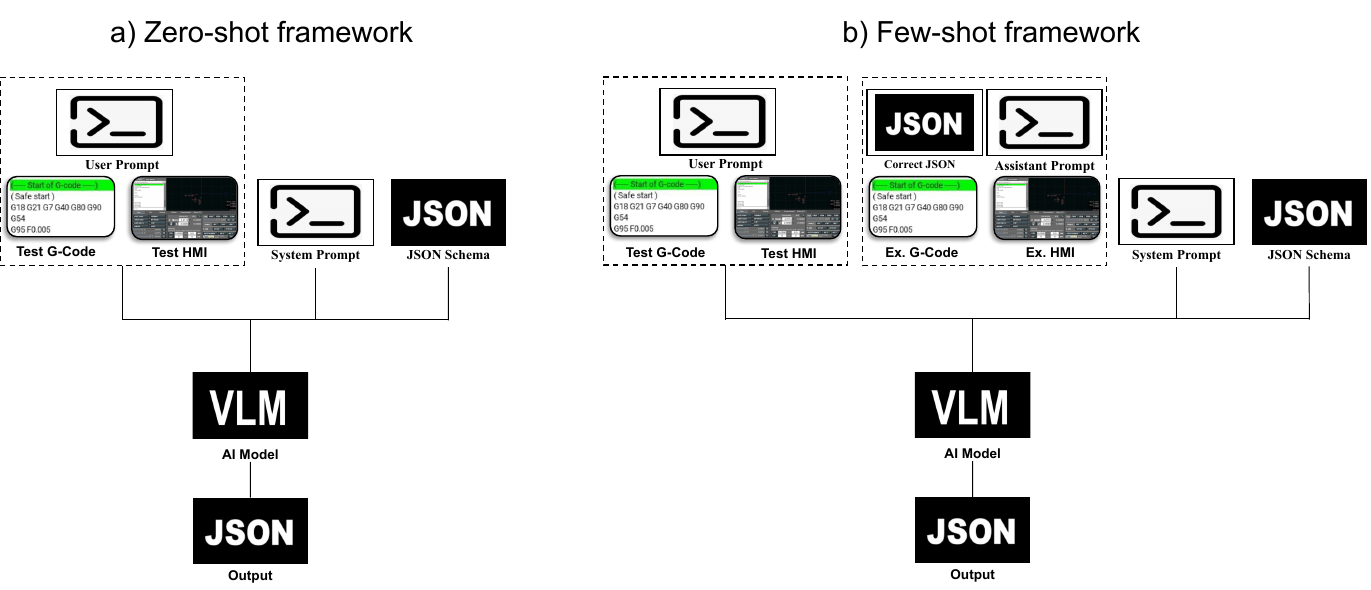}
    \caption{(a) In the zero-shot setting, the VLM initial instructions (system prompt), the specific request (user prompt), and the required output format (JSON schema).(b) The few-shot setting includes an extra assistant message that includes example HMI screenshots, along with their matching G-code and (correct) JSON outputs.}
    \label{fig:method}
\end{figure*}

This study introduces a VLM-based verification methodology that integrates two complementary modalities: textual G-code commands and their associated machine status displayed on HMI presented in Figure \ref{fig:HMI}. The framework aims to verify the syntactic accuracy of the G-code, retrieve machine-state data from the HMI display, and assess their alignment to ensure consistency between instruction and execution. On the HMI display, three primary status indicators, \textit{COLLET CLAMPED}, \textit{REF X}, and \textit{REF Z}, represent the machine’s operational readiness state. The analysis specifically focuses on these indicators to confirm the model’s correct recognition of visual cues. The methodology is organized into three subsequent sections. First, the multimodal input integration, which connects two types of information streams, is described. Next, the prompting strategy to guide the model’s communication is presented. Figure \ref{fig:method} illustrates the model framework from input integration to structured output generation.


\subsection{Multimodal Input Integration}

The first step is to prepare the multimodal G-code and HMI input pairs. The text input is imported in its original form without syntax preprocessing, and the HMI screenshot is loaded as an RGB image to ensure correct color-based interpretation. An optional preprocessing step identifies a cropped section containing the key indicators within the HMI screenshot. The model may receive the complete interface screenshot, only the right-hand control panel cluster, or a combination of these inputs. The crop is defined using percentage-based bounding boxes. These options allowed the model to be configures with different display setups and identify either on the whole interface or only the most important visual cues.

Following input preparation as a structured JSON schema shown Table \ref{tab:json_schema_summary} was constructed to define the validation fields for both data modalities. This schema outlines the structure through which the reasoning and output is reported by the model in a machine readable format. The mentioned structure contains: \textit{slots (collet_clamped, refx, and refz)} that represent boolean values derived from interface key indicators Specifically, when the status lights are active and green, the value is set to True or otherwise False. Any false indicator violates the readiness of the machine which results are stored in \textit{HMI errors}.

\begin{table}[ht]
\centering
\caption{Structure of json schema for the output. \\
T = boolean, S = string, [ ] = array.}
\renewcommand{\arraystretch}{1.2}
\setlength{\tabcolsep}{8pt}
\small
\begin{tabular}{
  p{0.25\linewidth}
  >{\hspace{0.85cm}}p{0.60\linewidth}
}
\toprule
\textbf{Field} & \textbf{Content / Type} \\
\midrule

\texttt{slots} &
\begin{minipage}[t]{\linewidth}
\ttfamily
\{collet\_clamped:T,\\
\hspace*{1em}refx:T,\\
\hspace*{1em}refz:T,\\
\hspace*{1em}HMI issues:[S]\}\\
\end{minipage} \\[6pt]

\texttt{gcode\_validity} &
\begin{minipage}[t]{\linewidth}
\ttfamily
\{valid:T,\\
\hspace*{1em}g-code errors:[S]\}\\
\end{minipage} \\[6pt]

\texttt{HMI and G-code compliance} &
\begin{minipage}[t]{\linewidth}
\ttfamily
\{consistent:T,\\
\hspace*{1em}HMI and G-code errors:[S]\}\\
\end{minipage} \\[6pt]

\texttt{corrections} &
\begin{minipage}[t]{\linewidth}
\ttfamily
     [S] 
\end{minipage} \\

\bottomrule
\end{tabular}
\label{tab:json_schema_summary}
\end{table}

In addition to slot states, \textit{gcode_validity} is a structural checkpoint. It stores a boolean value indicating if the G-code is properly structured and logs any detected \textit{G-code errors}. Furthermore, \textit{HMI and G-code compliance} evaluates the overall consistency based on the rules specified in the system prompt and reports any identified \textit{HMI and G-code errors}. Finally, an additional field, \textit{corrections}, generates explanatory notes and corrective feedback based on previous fields. Every identified discrepancy is reported in natural language to allow interpretable and field-level verification.

\subsection{Prompting Strategy}

Following input integration and setting the structural schema for the output, the next stage of the methodology is implemented through prompt-based reasoning without any model fine-tuning. All verification tasks were conducted using GPT-4.1, a VLM capable of analyzing both image and text, with a fixed temperature at zero to ensure deterministic outputs across the runs.

The framework uses three types of prompts to guide the model’s behavior during inference:

\begin{enumerate}
    \item The system prompt that defines the model's identity, role, verification setting, instructions about navigating inputs, compliance, and reporting rules. This prompt also embeds deterministic compliance logic that governs how textual and visual information are interpreted. Specifically, spindle control commands mandate \textit{collet_clamped = true}, any motion commands with X require \textit{refx = true}, and similarly with Z. These mappings are hard-coded within the system instructions and are applied consistently across all inferences. 
    \item The user prompt \textit{"Inspect the attached HMI image and G-code using the system rules. output only one JSON object. Check three LEDs: COLLET CLAMPED, REF X, REF Z. If any LED is dark or unclear, set false and include the exact issue line. Ignore G-code spacing."} that delivers the test input and has a fixed instruction enforcing a JSON-only response.
    \item The assistant prompt that is a teaching note and instructs the model by showing complete multimodal inputs as examples and their correct JSON file to illustrate the correct mapping between inputs and outputs.
\end{enumerate}

The method consists of two separate tasks: 1) a zero-shot prompting strategy and 2) a few-shot prompting strategy. In both configurations, the input is composed of three main components: (1) the system prompt (2) The structured JSON schema that will be used to generate the updated JSON file; (3) the test-set inputs consisting of the G-code, the HMI image, and a user prompt explaining the task. In the few-shot setting, an additional set of examples that includes correct and incorrect G-code segments paired with corresponding HMI images, an assistant prompt illustrating examples, and a correct JSON file, is provided to the model to support contextual learning from observed patterns. Within each inference, the G-code is parsed according to predefined rules encoded directly in instruction section of the system prompt. The model detects invalid commands, improperly formatted numbers, or conflicting operational modes as faulty, setting \textit{valid} flag to false and appending explanations in \textit{error} field. Once inference is complete, the generated results are automatically validated using JSON schema to maintain structural consistency and machine-readable results to enable reliable evaluation in subsequent stages.


Prior to inclusion, all example responses are verified according to the schema to ensure prompt consistency and avoid format deviations.

\section{Experimental Setup}
\label{sec:experiments}
This section describes the data collection procedure, followed by the implementation details and the evaluation method used to assess the framework.

\subsection{Data Collection}
To evaluate how well the model could verify machining states using G-code and HMI screenshot pairs, a dataset was constructed. The dataset was composed of sixteen unique G-Code and screenshot combinations distributed across eight experimental scenarios, each with two instances. In every instance, a G-Code text file along with its associated HMI screenshot was obtained from the PathPilot Lathe 15-L Slant Pro interface. The dataset was designed to balance operationally valid commands and machine states with faulty ones.

\begin{figure}[h!]
    \centering
    \includegraphics[width=0.85\linewidth, keepaspectratio]{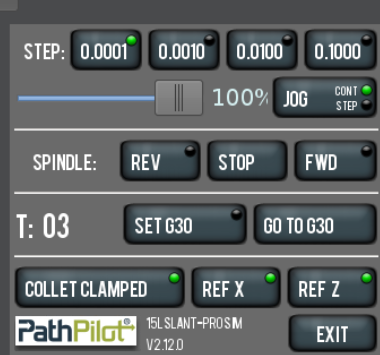}
    \caption{Example of the right-hand cluster crop from HMI interface. The screenshot shows the COLLET CLAMP, REF X, and REF Z are all illuminated green.}
    \label{fig:cluster}
\end{figure}
As a part of a multimodal dataset, the visual component, collected as screenshots from PathPilot, displays essential state indicators: clamp status, X-axis, and Z-axis reference. Within each screenshot, machine status and readiness are explicitly signaled by the indicator located at the edge of these signals as shown in Figure \ref{fig:cluster}. Specifically, when the collet is clamped, it secures the workpiece, and its corresponding indicator is illuminated green. A same condition is applied for homing: when the X and Z axis are referenced the tool is positioned at the home location along both axis. As a result, each screenshot reflects a specific combination of these three indicators representing a context that may occur during CNC operation. This variation enables the evaluation of how well the model can distinguish between active and inactive states and assess consistency between G-Code and HMI machine state.

To provide an overview of this framework, Table \ref{tab:cnc_gcode_syntax} lists the instance composition, indicator states, and G-Code errors. Building upon this visual foundation, each instance defines a specific operational scenario designed to test the model's ability to depict and represent machine states accurately. For instance, Scenario 1 involves contradictory motion commands (G00 and G01 in the same block) along with X and Z motions, whereas Scenario 2 includes invalid commands and spindle initiation with open collet. In Scenarios 3 and 4, a mix of correct and erroneous codes is tested under partially referenced states to observe how the model discriminates between syntactically valid structures and improper machine states. Subsequently, Scenarios 5 and 6 address incomplete motion codes, while Scenarios 7 and 8 focus more on the compliance state without clamping and referencing. 

\begin{table}[h!]
\centering
\caption{CNC HMI–G-code syntactic validation across eight scenarios.
T=True, F=False; C=Collet, RX=REF X, RZ=REF Z.}
\label{tab:cnc_gcode_syntax}

\begin{tabularx}{\linewidth}{c c c >{\raggedright\arraybackslash}X}
\toprule
Scenario(Inst) & Indicator (C/RX/RZ) & Val & GErr \\
\midrule
S1(i1) & T/F/F & F & Modal conflict \\
S1(i2) & T/F/F & T & -- \\
S2(i1) & F/F/F & F & Invalid command \\
S2(i2) & T/F/T & T & -- \\
S3(i1) & T/T/F & F & multiple errors \\
S3(i2) & T/T/F & T & -- \\
S4(i1) & T/F/T & F & Non-numeric coordinate \\
S4(i2) & T/F/T & T & -- \\
S5(i1) & F/T/F & F & Feed F missing value \\
S5(i2) & F/T/F & T & -- \\
S6(i1) & F/T/T & F & Unknown code \\
S6(i2) & F/T/T & T & -- \\
S7(i1) & T/T/T & F & Empty motion block \\
S7(i2) & T/T/T & T & -- \\
S8(i1) & F/F/F & F & Unsafe feed \\
S8(i2) & F/F/F & T & -- \\
\bottomrule
\end{tabularx}
\end{table}


Following the construction of the dataset, a seven-example few-shot framework was developed to guide the model’s reasoning across multiple modalities. By familiarizing the model with a limited but diverse set of examples before testing the new instances, the few-shot configuration enables it to extend the reasoning to new cases. These examples demonstrate expected verification outcomes, including individual G-code syntax errors, motion violations, proper spindle activation and axis referencing, as well as a fully valid configuration. For each example, a correct JSON output was provided to expose the model to a structured reference pattern for the reasoning flow. Furthermore, this design enables the comparison between the few-shot and zero-shot conditions.

\subsection{Implementation Procedure}

In order to prevent data leakage, test sets, examples, and prompts need to be introduced to the model by an ordered sequence. Each test instance starts by loading the G-code text file and corresponding interface screenshot captured from the HMI. When a specific configuration is specified, the system extracts the right-hand cluster with the percentage-based bounding boxes. By standardizing the input layout, this step ensures that visual data supplied to the model remain spatially consistent across all test cases. As a verification measure, an optional debug overlay highlights the extracted regions to confirm accurate extraction. In few-shot executions, example datasets that are synchronized with the schema are used during inference to provide in-context references.

During the inference, these multimodal inputs are organized into a structured message sequence constructed to resemble model's reasoning process. The conversation sequence starts with a system prompt specifying the verification policies and checking guidelines. These instructions are optionally followed by a brief few-shot instructional guide alongside the set of examples. In the final step, the user message provides the test G-code and a related HMI screenshot, following the same multimodal structure established in the examples.

Following the completion of inference, the  model response is validated against the schema which allows comparison and evaluation, and this is then stored. Batch runs automate the experiments and expand this workflow over multiple test scenario groups to perform both few-shot and zero-shot settings under \textit{full} and \textit{full plus cluster} configurations. The \textit{full} mode uses the complete-view interface, while the \textit{full plus cluster} setup additionally incorporates the right-hand control panel area. 

\subsection{Evaluation Procedure}
\begin{figure}[h!]
    \centering
    \includegraphics[width=1\linewidth]{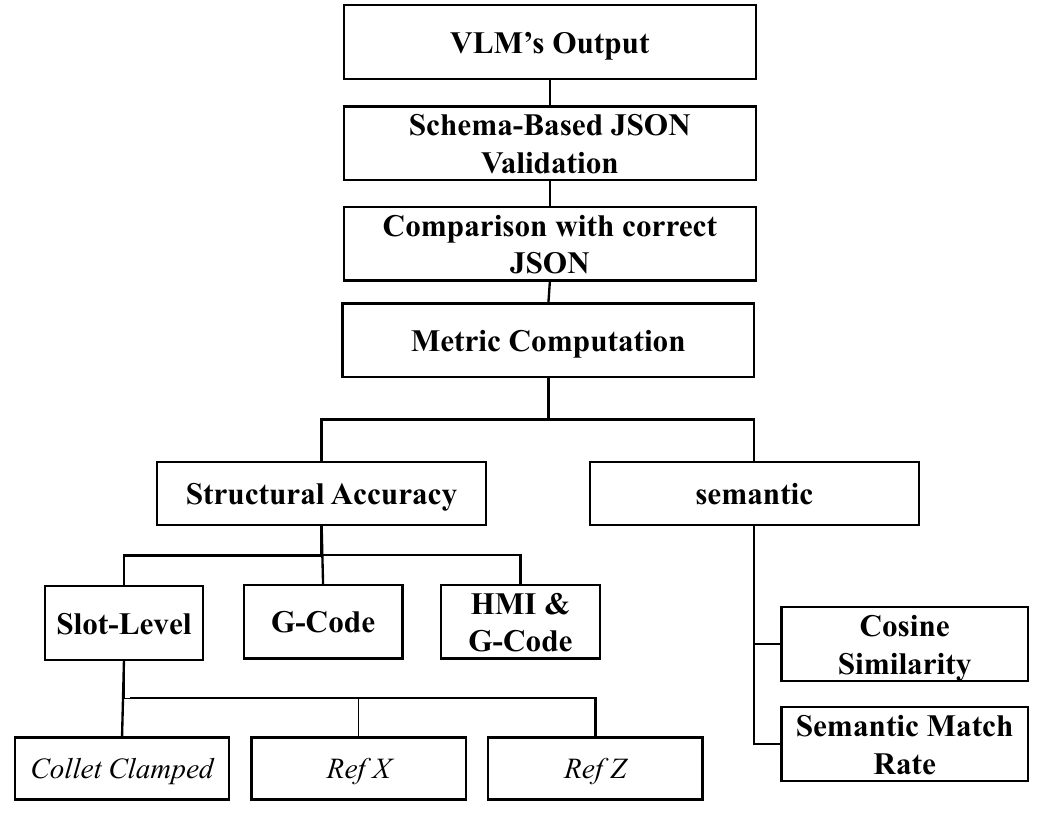}
    \caption{The figure outlines the evaluation steps applied to each instance's output.}
    \label{fig:evaluation}
\end{figure}

The evaluation process used a systematic procedure to ensure capturing the model's performance across all the scenarios, illustrated in Figure \ref{fig:evaluation}. Within each instance setup, the model generations were compared against the correct JSON schema of that instance. A schema validity was calculated to flag the outputs that were not generated according to the input JSON schema. Then, the comparison process was operated on two levels: structural accuracy and semantic alignment. The initial level of the evaluation first checks if the generated JSON output's fields match with the provided JSON schema in the input. Then comparison of each model-predicted slot (COLLET CLAMPED, REF X, and REF Z) to its corresponding booleans in the ground truth are evaluated. This step also assesses whether the prediction determines the validity of the G-code boolean and whether the compliance of the HMI and G-code matches the ground truth. After this comparison, accuracy for each HMI slot, G-code validity, and compliance accuracy is calculated. 

In addition to this structural binary validation, a semantic similarity-based assessment was implemented to evaluate natural language textual outputs, including HMI errors, G-code errors, combined HMI and G-code errors, and corrections. As a result, using the \textit{OpenAI text-embedding-3-small} model, cosine similarity was calculated between the model-generated and correct text vectors. When cosine similarity exceeded 0.80, the pair was considered meaning equivalent, which contributes to calculations of the average match rate (ratio of aligned items) and average similarity of matches (mean cosine similarity among aligned pairs).  Collectively, these steps generate interpretable schema-bounded verification outputs to enable error diagnosis and systematic evaluation.

\section{Results}
\label{sec:results}

To assess the effectiveness of the proposed framework, this study analyzed the findings in terms of structural metrics that evaluate binary outcome field accuracies and semantic alignment based on text fields, based on match rate and cosine similarity. All analyses were conducted across four distinct configurations: Zero-shot (ZS) with the full image (Full), ZS with the added cluster crop (+Clust), Few-shot (FS) with Full, and FS with +Clust.

As indicated by Table \ref{tab:struct} throughout the structural assessment, schema validity remained completely accurate across all four configurations. This perfect schema validity suggests that the generated outputs remained machine-readable and ready for later evaluation stages. Within this structural readability, G-code validity shows a steady and strong performance in ZS +Clust and both FS modes. Notably, G-code and HMI compliance accuracy increases from 0.625 to 0.750 with FS prompting. Overall, these findings imply that FS prompting enhances the model's reasoning.

\begin{figure}[h!]
    \centering
    \includegraphics[width=1\linewidth, keepaspectratio]{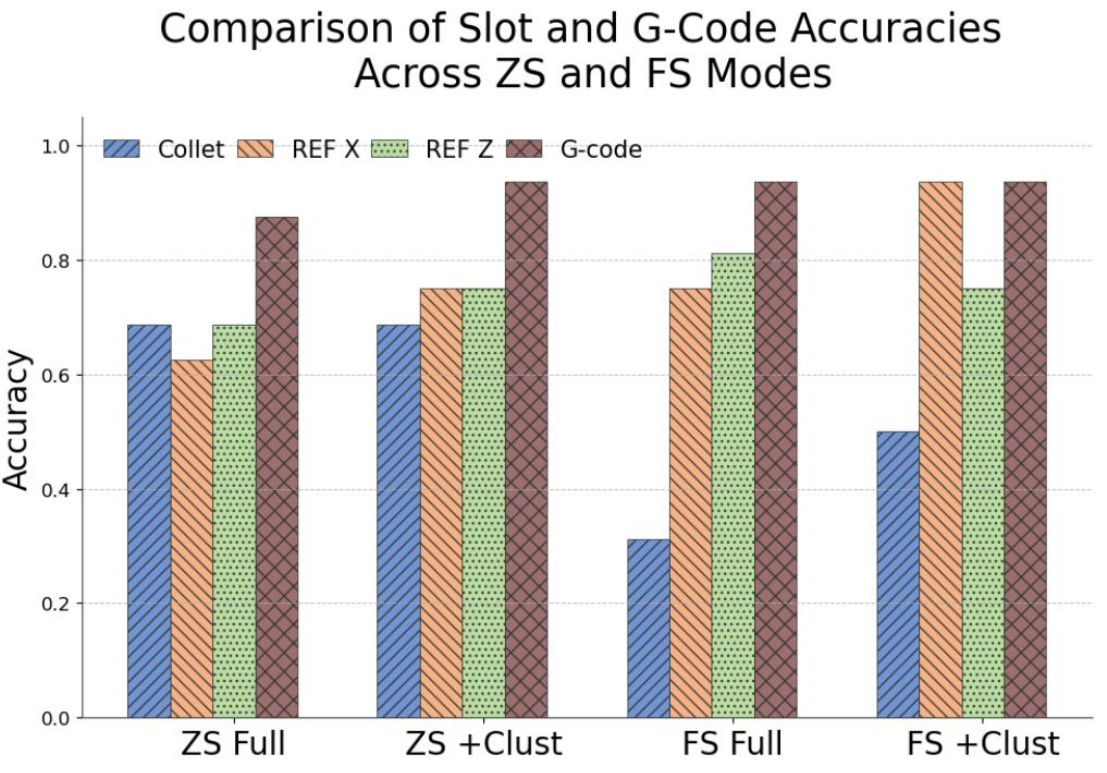}
    \caption{This figure summarizes slot-level structural accuracies comparing zero-shot (ZS) and few-shot (FS) configurations under full and cluster visual inputs.}
    \label{fig:accuracy}
\end{figure}

Examining the individual slot boolean, illustrated in Figure \ref{fig:accuracy} results reveals that the model's performance varies across COLLET CLAMPED, Ref X, and Ref Z indicators. The Ref X indicator shows the most significant increase when +Clust accompanies FS examples. Specifically, its accuracy rose from 0.625 in the ZS Full to 0.75 when the cluster was added (ZS +Clust) and reached 0.938 when both FS examples and +Clust were used together. In contrast, the Ref Z indicator followed a different pattern. The model prediction showed that it performed highest under an FS Full configuration, then showed a minor decrease after the cluster was added. Thus, this observation suggests that adding a cluster contributes more effectively to Ref X than Z, particularly in this dataset. Meanwhile, COLLET CLAMPED indicator exhibited yet another different pattern. It reached its peak accuracy in ZS configurations and drops under an FS configuration. 

\begin{table}[hbt!]
\centering
\small
\setlength{\tabcolsep}{4pt}
\renewcommand{\arraystretch}{1.15}
\caption{Structural metrics: Zero-shot vs Few-shot.}
\label{tab:struct}
\begin{tabular}{lcccc}
\toprule
\textbf{Metric} & \textbf{ZS Full} & \textbf{ZS +Clust} & \textbf{FS Full} & \textbf{FS +Clust} \\
\midrule
Schema Validity            & 1.000 & 1.000 & 1.000 & 1.000 \\
Collet clamped             & \textbf{0.688} & \textbf{0.688} & 0.312 & 0.500 \\
Ref X                      & 0.625 & 0.750 & 0.750 & \textbf{0.938} \\
Ref Z                      & 0.688 & 0.750 & \textbf{0.812} & 0.750 \\
G-code Validity Accuracy   & 0.875 & \textbf{0.938} & \textbf{0.938} & \textbf{0.938} \\
Compliance Accuracy        & 0.688 & 0.625 & \textbf{0.750} & \textbf{0.750} \\
\bottomrule
\end{tabular}
\end{table}

Collectively, these results suggest that combining FS examples with cluster crops primarily improves the model's ability to detect the indicator. However, the lack of improvement for Collet Clamped may stem from its visually different characteristics and size compared to the other two indicators. In G-Code validity, accuracy is consistent beyond the ZS setup, suggesting that the metrics' dependency lies in textual input rather than configuration. Therefore, the transition from full-image to clustered cropping complements the transition from ZS to FS, as cropping helps strengthen the FS reasoning of the model.

Complementing the structural analysis slot-level metrics, semantic analysis for each G-Code Error, HMI Error, and their combined category aims to corroborate the findings from the structural analysis in natural language. The results indicate how FS exemplars standardize the language use without distorting the information, as reflected by both the match-rate shown in Table \ref{tab:sem_match} and cosine-similarity scores presented in Table \ref{tab:sem_cos}. The most visible enhancements are observed among G-code and combined HMI and G-code error, where both cosine similarity and match rate increase significantly compared to ZS runs.
Similarly, HMI related errors show a moderate but stable enhancement, showing improved linguistic clarity and consistency, matching the trends in structural metrics. However, the corrections output demonstrates that the FS configuration improves phrasing consistency even in open-ended generative responses.

\begin{table}[hbt!]
\centering
\small
\setlength{\tabcolsep}{4pt}
\renewcommand{\arraystretch}{1.15}
\caption{Semantic cosine similarity: Zero-shot vs Few-shot.}
\label{tab:sem_cos}
\begin{tabular}{lcccc}
\toprule
\textbf{Category} & \textbf{ZS Full} & \textbf{ZS +Clust} & \textbf{FS Full} & \textbf{FS +Clust} \\
\midrule
G-Code Error       & 0.618 & 0.616 & \textbf{0.771} & 0.725 \\
HMI Error      & 0.438 & 0.375 & \textbf{0.740} & 0.688 \\
HMI and G-Code Error   & 0.500 & 0.438 & 0.688 & \textbf{0.750} \\

Corrections  & 0.385 & 0.276 & \textbf{0.680} & \textbf{0.680} \\
\bottomrule
\end{tabular}
\end{table}

\begin{table}[hbt!]
\centering
\small
\setlength{\tabcolsep}{4pt}
\renewcommand{\arraystretch}{1.15}
\caption{Semantic match rate: Zero-shot vs Few-shot.}
\label{tab:sem_match}
\begin{tabular}{lcccc}
\toprule
\textbf{Category} & \textbf{ZS Full} & \textbf{ZS +Clust} & \textbf{FS Full} & \textbf{FS +Clust} \\
\midrule
G-Code Error       & 0.594 & 0.594 & \textbf{0.781} & 0.719 \\
HMI Error       & 0.292 & 0.276 & 0.448 & \textbf{0.474} \\
HMI and G-Code Error   & 0.396 & 0.385 & 0.479 & \textbf{0.583} \\

Corrections  & 0.292 & 0.276 & 0.438 & \textbf{0.479} \\
\bottomrule
\end{tabular}
\end{table}


Combining two sets of findings, structural and semantic, uncovers a complementary relationship between FS prompting and adding visual crops. Through enhancing linguistic normalization and alignment with formatting standards, FS prompting strengthens reasoning and readability in the compliance performance and open-text fields in corrections and HMI-related descriptions. Alternatively, adding the right-hand cluster to the input mainly improves the recognition of localized machine state indicators, most notably Ref X, while contributing even further to the enhancement by adding the exemplars. The balance between the two approaches, namely adding examples and adding crops, clarifies why FS+cluster achieves the highest Ref X accuracy and strongest cross-modal correspondence while FS performs better in general for text-based analysis and G-code validity. Overall, the results indicate that the configuration selection can be adjusted according to the task objectives.



\section{Conclusions}
\label{sec:conclusions}
This study introduces a multimodal framework for the simultaneous verification of G-code and HMI states in CNC machining. The proposed approach addresses the current gap in LLM-based frameworks, which lack multimodal understanding to interpret visual indicators of machine state. This was achieved by integrating paired G-codes and HMI screenshots from the PathPilot Lathe 15-L Slant Pro as textual and visual inputs to a vision-language model (GPT-4.1). These data were organized into eight scenarios, each with two instances, and processed within a JSON schema to ensure interpretability. The framework was evaluated and tested under ZS and FS prompting conditions, with alternative views of the interface display, including full and clustered crops of the control panel. 

The findings emphasize that FS prompting led to overall enhancement of G-code and HMI compliance accuracy, whereas G-code validity remained consistent and stable beyond ZS Full configuration. Additionally, the state indicator slots, Ref X, achieved the highest accuracy under FS +Clust for 0.938, while Ref Z performed best under FS Full mode. In contrast, collet clamped remained most accurate under ZS configuration. In terms of semantic alignment, cosine similarity and match rate, increases were observed across all natural language error fields under FS prompting. Among these fields, correction (match rate up to 0.479; similarity up to 0.680) and HMI-aware content (including HMI Error similarity up to 0.740) had the most substantial increase. These results suggest that FS prompting promotes a more stable language generation, while cropping the interface enhances the local visual cues that matter for axis references. Overall, the results suggest that multimodal reasoning improves a safer G-code verification while supporting a more transparent human-machine interaction.
The key limitation of this study is the task-specific dataset which restricts broader generalization to other machine types and unseen operating conditions. Moreover, this framework is sensitive to the design of the prompt. Therefore, making stable prompts is essential as minor adjustments in the phrasing of the prompt can result in different reasoning paths. Future extensions can address these limitations by expanding dataset size and scenario diversity, incorporating sensor feedback, and exploring prompt standardization strategies to enhance reliability and generalization in practical manufacturing environments.




\nocite{*}

\bibliographystyle{asmeconf}  
\bibliography{asmeconf-sample}

\end{document}